\documentclass[11pt]{article}
\usepackage{acl2015}
\usepackage{times}
\usepackage{url}
\usepackage{graphicx}
\usepackage{latexsym}
\def\BibTeX{{\rm B\kern-.05em{\sc i\kern-.025em b}\kern-.08em
    T\kern-.1667em\lower.7ex\hbox{E}\kern-.125emX}}

\title{Machine Translation : From Statistical to modern Deep-learning practices}

\author{Siddhant Srivastava \\
  Soft Computing Laboratory \\
  ABV-IIITM Gwalior \\
  {\tt siddhant.srivastava11@gmail.com} \\\And
  Anupam Shukla \\
  IIIT Pune \\
  {\tt dranupamshukla@gmail.com} \\\And
  Ritu Tiwari \\
  Soft Computing Laboratory \\
  ABV-IIITM Gwalior \\
  {\tt tiwariritu2@gmail.com}}

\date{}

\begin{document}
\maketitle
\begin{abstract}
  Machine translation (MT) is an area of study in Natural Language processing which deals with the automatic translation of human language, from one language to another by the computer. Having a rich research history spanning nearly three decades, Machine translation is one of the most sought after area of research in the linguistics and computational community. In this paper, we investigate the models based on deep learning that have achieved substantial progress in recent years and becoming the prominent method in MT. We shall discuss the two main deep-learning based Machine Translation methods, one at \textit{component or domain level} which leverages deep learning models to enhance the efficacy of Statistical Machine Translation (SMT) and \textit{end-to-end deep learning models in MT} which uses neural networks to find correspondence between the source and target languages using the \textit{encoder-decoder architecture}. We conclude this paper by providing a time line of the major research problems solved by the researchers and also provide a comprehensive overview of present areas of research in Neural Machine Translation.
\end{abstract}

\section{Introduction}
Machine Translation, which is a field of study under natural language processing, targets at translating natural language automatically using machines. Data-driven machine translation has become the dominant field of study due to the accessibility of large parallel corpora. The main objective of data-driven machine translation is to translate unseen source language, given that the systems learn translation knowledge from sentence aligned bi-lingual training data.

Statistical Machine Translation (SMT) is a data-driven approach which uses probabilistic models to capture the translation process. Early models in SMT were based on generative models taking a word as the basic entity (Brown 1993), maximum entropy based discriminative models using features learned from sentences (Och 2002), simple and hierarchical phrases (Koehn 2003, Chiang 2007). These methods have been greatly used since 2002 despite the fact that discriminative models faced with the challenge of data sparsity. Discrete word based representations made SMT susceptible to learning poor estimate on the account of low count events. Also, designing features for SMT manually is a difficult task and require domain language, which is hard keeping in mind the variety and complexity of different natural languages.

Later years have witnessed the exceptional success of deep learning applications in machine translation. Deep learning approaches have outstripped statistical methods in almost all sub-fields of MT and have become the \textit{de facto} method in both academia as well as in the industry. In this paper, we will discuss the two domains where deep learning has been greatly employed in MT. We will briefly discuss \textit{Component or Domain wise deep learning methods for machine translation} (Devline 2014) which uses deep learning models to improve the effectiveness of different components used in SMT including language models, transition models, and re-structuring models. Our main focus in on \textit{end-to-end deep learning models for machine translation} (Sutskever 2014, Bahdanau 2014) that uses neural networks to extract correspondence between a source and target language directly in a holistic manner without using any hand-crafted features. These models are now recognised as \textit{Neural Machine translation} (NMT).

The paper is arranged as follows. We will first introduce the basic definitions and objective in Machine Translation. The next section will have a brief discussion on component-wise deep learning models and how they improve SMT based models. Next, we'll be focusing on end-to-end Neural Machine Translation mentioning the challenges faced in these models, we will discuss the currently employed encoder-decoder model focusing mainly on network architecture used in the paradigm which is the current area of research. We will conclude by giving the time-line of major breakthroughs in NMT and will also propose the future research areas in neural architecture development.

\section{Machine translation : Essentials}

Let $x$ denote the source language and $y$ denote the target language, given a set of model parameters $\theta$ , the aim of any machine translation algorithm is to find the translation having maximum probability $\hat{y}$:
\begin{equation}
    \hat{y}=argmax_y  {P(y|x;\theta)}.
\end{equation}
The decision rule is re-written using Bayes' rule as (Brown 1993):
\begin{equation}
    \hat{y} = argmax_y {\frac{P(y;\theta_{lm})P(x|y;\theta_{tm})}{P(x)}}.
    \end{equation}
\begin{equation}
   \hat{y} = argmax_y {P(y;\theta_{lm})P(x|y;\theta_{tm})}. 
\end{equation}
Where $P(y;\theta_{lm})$ is called as \textit{language model}, and $P(x|y;\theta_{tm})$ is called as \textit{transition model}.
The translation model in addition, is defined as generative model, which is disintegrated via latent structures. 
\begin{equation}
    P(x|y;\theta_{tm}) = \sum_{z}P(x,y|z;\theta_{tm}).
\end{equation}
Where, $z$ denotes the latent structures like word alignment between source language and target language. The key problem with this approach is that it is hard to generalize because of dependencies among sub-models. To introduce knowledge sources in SMT, (Och 2002) uses log-linear models.
\begin{equation}
    P(y,x|\theta) = \frac{\sum_z exp(theta*\psi(x,y,z))}{\sum_y' \sum_z' exp(theta*\psi(x',y,z'))}
\end{equation}
Where $\psi(x,y,z)$ is a set of features defining the translation process and $\theta$ denotes corresponding weights for each feature. (Koen 2003) introduces phrase-based translational model which is widely used in academia and industry, the basic idea is to utilize phrases to capture word selection and restructuring the local context. The translation model in phrase-based Statistical Machine Translation is divided in three main steps or sub-models: (1) segmentation of source sentence to phrases. (2) transforming each source phrase to target phrase. (3) restructuring target phrases to match target language order. The addition of target phrases corresponds to target sentence. 

Statistical Machine Translation suffers from two problems, which are \textit{data sparsity} and \textit{feature engineering}. Due to discrete symbolic representation, the statistical model is disposed to learning weak estimates of model parameter on account of low numbers. As a result, conventional conventional SMT resort to using simple features instead of complex features which sets an unavoidable bar on the model's effectiveness. The second challenge faced by SMT is of feature engineering, usual practice in feature design of SMT includes annotating hand crafted features to capture local syntactic and semantic features. Since there can be millions of such features and mapping those features from one language to another can be a cumbersome task, designing general features in SMT remains a challenge. 

\section{Component or Domain wise Deep-learning methods in Statistical Machine Translation}

In recent years, deep learning based approaches have been greatly studies to mitigate the issues faced by SMT, data sparsity and feature engineering. In this section, we briefly discuss how deep learning based models have been successfully used in key components of SMT, Word alignment, transition rule probability estimation, phrase restructuring model and language model.

\subsection{Word Alignment}
The role of word alignment is to find correspondence between words present in parallel corpus (Brown 1993; Vogel 1996). In SMT, word alignment is taken as hidden variable in generative models. Word alignment is basically modelled as $P(x, z|y;\theta)$, The Hidden Markov model is the most widely used SMT alignment model (Vogel 1996) and the conventional training objective is to maximize the log-likelihood of training data. The drawback of using conventional generative model is that it fails to capture complex relationships among natural language due to data sparsity when using discrete symbolic representation.

(Yang 2013) were the first to propose context dependant deep neural network for word alignment, there idea was to capture contextual features using continuous word representations as input to a feed forward-neural network. (Tamura 2014) used Recurrent neural network (RNN) to calculate alignment score directly taking previous alignment scores as input, The authors reported better performance than the Yang's model.  

\subsection{Transition rule probability estimation}

In a phrase based SMT, we may get multiple transition rule from word aligned training data, the objective becomes to select the most apt rules during decoding phase. Usually, rule selection is done using transition probabilities and computed using MLE (Koehn 2003). The problem with this approach is that it suffers from data sparsity and fails to capture deep semantics and context. Deep learning techniques aim to alleviate these issues, (Gao 2014) calculate transition score among source and target phrase in a low dimensional vector space using feed forward network. (Devlin 2014) proposed a joint neural model which aimed at modelling both source and target context to predict transition score using feed forward network.

\subsection{Reordering phrases}

After scoring phrases between source and target sentence using transition scores, the next step is to order target phrases to produce a well-formed sentence. Earlier SMT computed phrase order taking discrete symbolic representations (Xiong 2016). (Li 2013; Li 2014) proposed a neural phrase reordering network which employed recursive autoencoders to learn continuous distributed representation of both source and target phrases, and making final ordering prediction with the help of feed forward network.

\subsection{Language Modelling}

Target phrases are combined to create target sentence or a larger partial translation in some cases, the role of language model is to determine that the larger translation is better than it's composed phrases. Conventional SMT's used n-gram language model to compute this conditional probability. Since n-gram model is count based, it suffered severely from data sparsity. Deep learning models helped alleviate this issue by computing conditional probability using continuous representation of words. (Bengio 2013) designed a feed forward network to digest n-gram model into a continuous vector space. (Vaswani 2013) integrated this neural network based n-gram model into the phrase decoding phase of SMT, he developed a recurrent neural network taking n-gram representations of fixed size. This model, using n-gram representation, assumed that the generation of current word depended on precursor $n-1$ words, this assumption was relaxed with using LSTM (Hochreichter 1997) and GRU (Cho 2014) based network which took into account all previous words to predict the current word.

\section{End-to-End Deep Learning for Machine translation}

End-to-End Machine Translation models also termed as Neural Machine Translation (NMT), aims to find a correspondence between source and target natural languages with the help of deep neural networks. The main difference between NMT and conventional Statistical Machine Translation (SMT) based approaches is that Neural model are capable of learning complex relationships among natural languages directly from the data, without resorting to manual hand features, which are hard to design.

The standard problem in Machine Translation remains the same, given a sequence of words in source language sentence $X = x_{1},....x_{j},....x_{J}$ and target language sentence $Y = y_{1},....y_{i},....y_{I}$, NMT tries to factor sentence level translation probability into context dependant sub-word translation probabilities.
\begin{equation}
    P(y|x;\theta) = \prod_{i=1}^{I} P(y_{i}|x,y_{<i};\theta)
\end{equation}
Here $y_{<i}$ is referred to as partial translation. There can be sparsity among context between source and target sentence when the sentences become too long, to solve this issue, (Sutskever 2014) proposed an encoder-decoder network which could represent variable length sentence to a fixed length vector representation and use this distributed vector to translate sentences.

\subsection{Encoder Decoder Framework for Machine Translation}

Neural Machine Translation models adhere to an Encoder-Decoder architecture, the role of encoder is to represent arbitrary length sentences to a fixed length real vector which is termed as context vector. This context vector contains all the necessary features which can be inferred from the source sentence itself. The decoder network takes this vector as input to output target sentence word by word. The ideal decoder is expected to output sentence which contains the full context of source language sentence.

Since source and target sentences are usually of different lengths, Initially (Sutskever 2014) proposed Recurrent Neural Network for both encoder and decoder networks, To address the problem of vanishing gradient and exploding gradients occurring due to dependencies among word pairs, Long Short Term Memory (LSTM) and Gated Recurrent Unit (GRU) were proposed instead of Vanilla RNN cell. Fig.1 Shows the architectural flow of a basic encoder-decoder based network.

Training in NMT is done by maximizing log-likelihood as the objective function:
\begin{equation}
    \hat{\theta} = argmax_\theta L(\theta)
\end{equation}
Where $L(\theta)$ is defined as:
\begin{equation}
    L(\theta) = \sum_{i=1}^{I} log P(y^{(i)}|x^{(i)};\theta)
\end{equation}
After training, learned parameters $\hat{\theta}$ is used for translation as: 
\begin{equation}
    \hat{y} = argmax_y P(y|x;\hat{\theta})
\end{equation}

\begin{figure} [t]
\includegraphics[width=8cm, height=7.5cm]{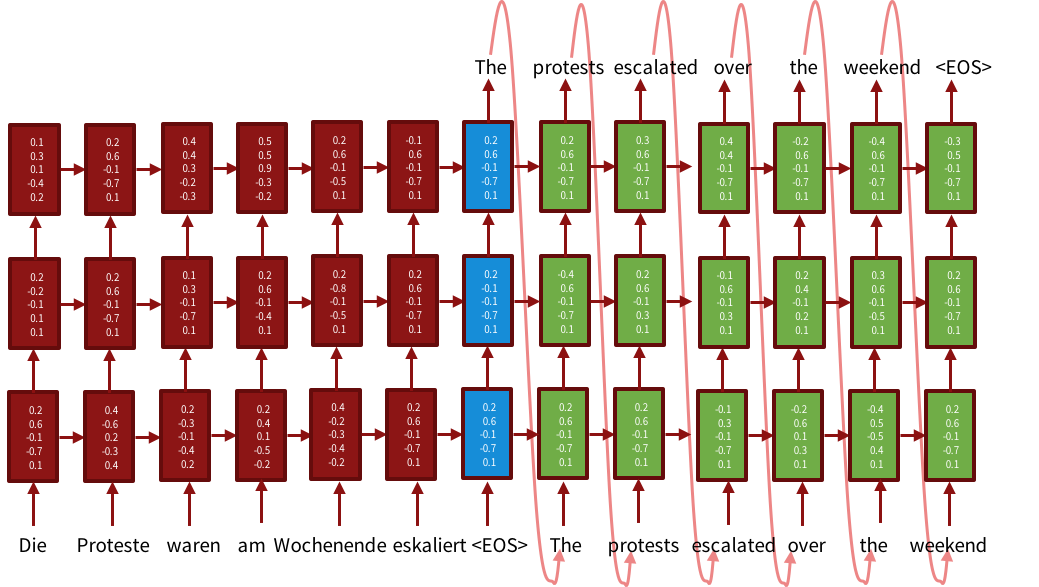}
\caption{Encoder-Decoder model for Machine Translation, Crimson boxes depict the hidden stated of encoder, Blue boxes shows "End of Sentence" EOS and Green boxes show hidden state of the decoder. credits (Neural Machine Translation - Tutorial ACL 2016)}
\label{fig:model}
\end{figure}
\subsection{Attention Mechanism in Neural Machine Translation}

The Encoder network proposed by (Sutskever 2014) represented source language sentence into a fixed length vector which was subsequently utilised by the Decoder network, through empirical testing, it was observed that the quality of translation greatly depended on the size of source sentence and decreased significantly by increasing the sentence size. 

To address this issue, (Bahdanau 2014) proposed to integrate an Attention mechanism inside the Encoder network and showed that this could dynamically select relevant portions of context in source sentence to produce target sentence. They used Bi-directional RNN (BRNN's) to capture global contexts:
\begin{equation}
    \overrightarrow{h_{s}} = f(x_{(s)},\overrightarrow{h_{s-1}},\theta)
\end{equation}

\begin{equation}
    \overleftarrow{h_{s}} = f(x_{(s)},\overleftarrow{h_{s-1}},\theta)
\end{equation}
The forward hidden state $\overrightarrow{h_{s}}$ and backward hidden state $\overleftarrow{h_{s}}$ are concatenated to capture sentence level context.
\begin{equation}
    h_{s} = [\overrightarrow{h_{s-1}};\overleftarrow{h_{s-1}}]
\end{equation}

The basic Ideology behind computing \textit{attention} is to find relevant portions in source text in order to generate target words in text, this is performed by computing attention weights first.
\begin{equation}
    \alpha_{j,i} = \frac{exp(a(t_{j-1},h_{i},\theta))}{\sum_{i'=1}^{I+1}exp(a(t_{j-1},h_{i'},\theta))}
\end{equation}

Where $a(t_{j-1},h_{i},\theta)$ is the alignment function which evaluates how well inputs are aligned with respect to position $i$ and output at position $i$. Context vector $c_{j}$ is computed as a weighted sum of hidden states of the source.

\begin{equation}
    c_{j} = \sum_{i=1}^{I+1} \alpha_{j,i}h_{i}
\end{equation}
And target hidden state is computed as follows. 
\begin{equation}
    t_{j} = f(y_{j-1},s_{j-1},c_{j},\theta)
\end{equation}
The difference between attention based NMT from original encoder-decoder based architecture is the way source context is computed, in original encoder-decoder, the source's hidden state is used to initialize target's initial hidden state whereas in attention mechanism, a weighted sum of hidden state is used which makes sure that the relevance of each and every source word in the sentence is well preserved in the context. This greatly improves the performance of translation and thus this has become the $state of the art$ model in neural machine translation. Fig.2 shows the architectural flow of attention based encoder and how this information is carried forward and utilized bu the decoder system.

\begin{figure} [t]
\includegraphics[width=8cm, height=6cm]{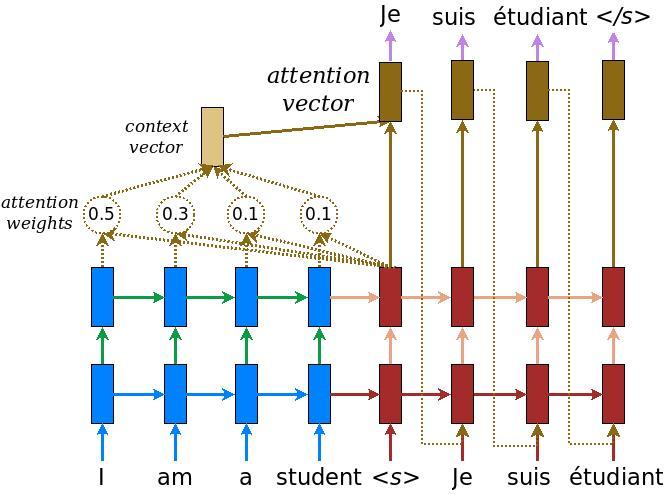}
\caption{Attention based Encoder-Decoder Architecture for Machine Translation. Most of the Architecture is similar to basic Encoder-Decoder with the addition of Context Vector Computed using attention weights for each word token, Attention vector is calculated using Context vector and hidden state of encoder. credits (Attention-based Neural Machine Translation with Keras, blog by Sigrid Keydana)}
\label{fig:model}
\end{figure}

\section{Challenges in Neural Machine Translation}
In this section, we will discuss some of the issues researchers faced in NMT and the solutions proposed by them. This section will include issues related (1) Inefficient performance due to large vocabulary, (2) Evaluation metric for end-to-end training in NMT,  (3) NMT training in low-quality data and (4) Network Architecture in NMT. Our main focus will be based on the discussion of Network architectures that have been proposed in recent years. Our main objective in this section is to pin point the current open ended research problems in the Neural Machine Translation and propose some of our own possible research areas.

\subsection{Inefficient performance due to large Vocabulary}
Because NMT uses word level tokens as input, the translation probability is calculated by normalizing over the total target words, the log likelihood of training data in turn depends over the computation of transition probabilities. Calculating the gradient of this log likelihood becomes an extraneous task for training NMT models as we need to enumerate through all the target words. 

For these reasons, (Sutskever 2014; Bahdanau 2014) trained their model on a subset of full vocabulary taking most frequent words for training and rest of the words as \textit{Out of Vocabulary (OOV)}, but this deteriorated the overall performance of the model significantly.

(Luong 2014) proposed a method to find correspondence between source and target OOV words and translate among the source and target as a separate step in pre-processing. One interesting approach to solve this problem is to use character level (Chung 2016; Luong 2016) and sub-word level (Sennrich 2015) tokens as input to the neural architecture. The intuition behind this approach is that using characters and sub-words greatly reduces the vocabulary making the computation of translation probability significantly faster.

\subsection{Evaluation Metric for end-to-end training in NMT}

The standard training metric used by all the machine translation systems is Maximum Likelihood Error (MLE), MLE finds the optimum model parameters by maximizing the log-likelihood of the training data. (Ranzato 2015) identified a likely drawback with this approach, He pointed that MLE works at word level for it's loss function whereas the evaluation metric used by Machine Translation is BLEU (Papineni 2002) and TER (Snover 2006) which are defined at the corpus or sentence level. The inconsistency between model training and evaluation poses problem for Neural Machine Translation.

To solve this issue (Shen 2015) proposed Minimum Risk Training (MRT) as a loss function while training Neural models, the idea behind this being that loss function should measure the difference between model predictions and the ground truth translation value, the optimal parameters should be computed by minimizing this loss. the training objective is defined as.
\begin{equation}
    \hat{\theta} = argmin_\theta {R(\theta)}
\end{equation}
\begin{equation}
    R(\theta) = \sum_{s=1}^{S} \sum_{y \in Y(x^{(s)})} P(y|x^{(s)};\theta)\Delta(y,y^{(s)})
\end{equation}
Where $Y(x^(s))$ is the set of all translations of $x^(s)$, $y$ and $y^(s)$ are the model predictions and ground truth respectively, $\Delta(y,y^{(s)})$ is the loss function computing the difference between prediction as ground truth values.
The advantage of MRT over MLE is that MRT can directly optimize model parameters with respect to evaluation metric. Also, MRT uses sentence level loss function, unlike word level in MLE and MRT is transparent to neural models and can be used in a number of artificial intelligence tasks. 

\subsection{NMT training in low-quality data}

NMT models owe their success to the parallel corpus containing parallel text as they are the main source of knowledge acquisition in terms of translation. As a result, the translation quality of NMT systems rely greatly on the quality and quantity of parallel corpora. NMT models have been effective in translating resource rich languages but for low resource language, the unavailability of large scale, high quality corpora poses a challenge for NMT as neural models learn poorly from low level counts. Results show that NMT performs worse than SMT in case of low data availability.

(Gulcehre 2015) proposed a solution by incorporating knowledge learned using mono-lingual data which is relatively abundant compared to parallel text, into the NMT model. They proposed to types of fusion, shallow fusion and deep fusion by incorporating the language model which is trained on mono-lingual data into the hidden state or the decoder of NMT. (Cheng 2016) proposed a semi-supervised learning approach to NMT by proposing a neural autoencoder for source-to-target and target-to-source which can be trained on both parallel and mono-lingual data. (Cheng 2017) proposed a pivot language based approach, the idea being that one can train an NMT by training it using source-to-pivot and pivot-to-target separately. 

\subsection{Neural Architectures for NMT}

Most of the encoder-decoder based NMT models have utilized RNN and It's variants LSTM and GRU. Recently, Convolution networks (CNN) and self attention networks have been studies and have produced promising results.

The issue with using Recurrent networks in NMT is that it works by serial computation and needs to maintain it's hidden step at each step of training. This makes the training highly inefficient and time consuming. (Gehring 2017) suggested that convolution networks can, in contrast, learn the fixed length hidden states using convolution operation. The main advantage of this approach being that convolution operation doesn't depend on previously computed values and can be parallelized for multi-core training. Also Convolution networks can be stacked one after the other to learn deeper context making it an ideal choice for both the encoder and decoder.

Recurrent networks compute dependency among words in a sentence in $O(n)$ whereas Convolution network can achieve the same in $O(log_{k}n)$ where $k$ is the size of convolution kernel.

(Vaswani 2017) proposed a model which could compute the dependency among each word pair in a sentence using only the attention layer stacked one after the other in both the encoder and decoder, he termed this as \textit{self-attention}. In their model, hidden state is computed using self-attention and feed forward network, they use positional encoding to introduce the feature based on the location of word in the sentence and their self-attention layer named as \textit{multi-head attention} is highly parallelizable. This model has shown to be highly parallelizable due to before mentioned reason and significantly speeds up NMT training, also resulting in better results than the baseline Recurrent network based models. Fig 3. shows the internal architecture of Transformer network as proposed by (Vaswani 2017), the figure shows encoder and decoder module separately.

\begin{figure} [t]
\includegraphics[width=8cm, height=12cm]{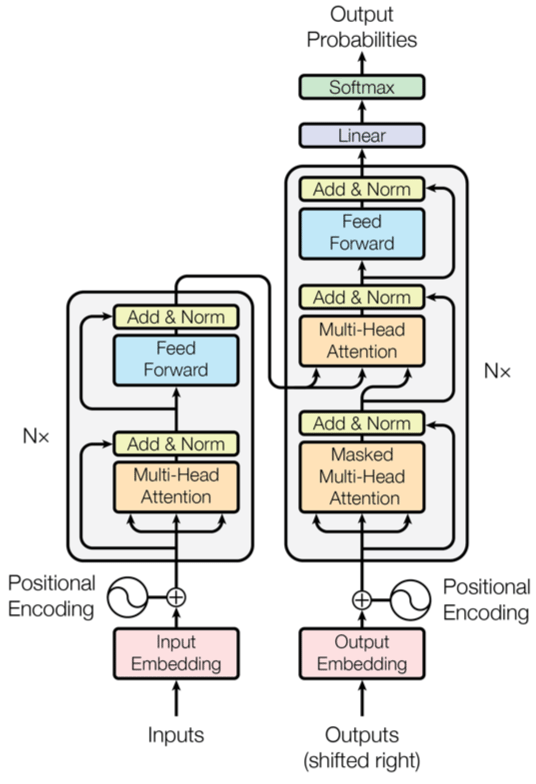}
\caption{Self-Attention Encoder-Decoder Transformer model. Encoder and Decoder both consists positional encoding and stacked layers of multi-head attention and feedforward network with the Decoder containing an additional Masked multi head attention. Transition Probabilities are calculated using linear layer followed by softmax. credits (Vaswani et al. 2017)}
\label{fig:model}
\end{figure}

Currently, there is no consensus regarding which neural architecture is the best and different architectures give different results depending on the problem in hand. Neural architecture is still considered to be the hottest and most active research field in Neural Machine Translation.

\section{Research gaps and open problems}

Deep learning methods have revolutionized the field of Machine Translation, with early efforts focusing on improving the key components of Statistical Machine Translation like word alignment (Yang 2013) , translation model, and phrase reordering and language model. Since 2010, most of the research has been shifted towards developing end-to-end neural models that could mitigate the need of extensive feature engineering . Neural models have successfully replaced Statistical models since their inception in all academic and industrial application.

Although Deep learning has accelerated research in Machine Translation community but nonetheless, Current NMT models are not free from imperfections and has certain limitations. In this section, we will delineate some existing research problems in NMT, our aim is to guide researchers and scholars working in this field to get acquainted with these issues and work towards it for even faster development in the field. Table 1. shows the major contributions done in the field of machine translation as years progress, we can see the inception of neural network architectures for machine translation beginning from the year 2014 and most breakthroughs are observed using the same neural approach. 

\subsection{Neural models inspired by linguistic approaches}
End-to-End models have been termed as the \textit{de facto} model in Machine Translation, but it is hard to interpret the internal computation of neural networks which is often simply said to be the \textit{"Black Box"} approach. One possible area of research is to develop linguistically motivated neural models for better interpretability. It is hard to discern knowledge from hidden state of current neural networks and as a result it is equally difficult to incorporate prior knowledge which is symbolic in nature into continuous representation of these states (Ding 2017). 

\subsection{Light weight neural models for learning through sparse data}
Another major drawback for NMT is data scarcity, It is well understood that NMT models are data hungry and requires millions of training instances for giving best results. The problem arises when there is not enough parallel corpora present for most of the language pairs in the world. Thus building models that can learn decent representation using relatively smaller data set is an actively researched problem today. One similar issue is to develop one-to-many and many-to-many language models instead of one-to-one models. Researchers are not sure how to common knowledge using neural network from a linguistic perspective, as this knowledge will help develop multi-lingual translation models instead of one-to-one models used today.

\subsection{Multi-modal Neural Architectures for present data}
One more problem is to develop multi-modal language translation models. Almost all the work done has been based on textual data. Research on developing continuous representation merging text, speech and visual data to develop multi-model systems is in full swing. Also since there is limited or no multi-model parallel corpora present, development of such databases is also an interesting field to explore and can also benefit multi-modal neural architectures.

\subsection{Parallel and distributed algorithms for training neural models}
Finally, current neural architectures rely heavily extensive computation power for giving competent results. Although there is no compute and storage shortage in current scenario, but it would be more efficient to come up with light neural models of language translation. Also Recurrent models cannot be parallelized due to which it is hard to develop distributed systems for model training. Fortunately, recent developments, with the emergence of Convolution networks and self-attention Networks can be parallelized and thus distributed among different systems. But because they contain millions of interdependent parameters, it makes it hard to distribute them among loosely coupled systems. Thus developing light neural architectures meant to be distributed can be new probable frontier of NMT.

\section{Conclusion}

In this paper, we discussed about Machine translation. We started our discussion from a brief discussion on basic Machine translation objective and terminologies along with early Statistical approaches (SMT) pointing out different components in each SMT system. We then discussed the role of deep learning models in improving different components of SMT, Then we shifted our discussion on end-to-end neural machine translation (NMT). Our discussion was largely based on the basic encoder-decoder based NMT, attention based model. We finally listed the challenges in Neural Translation models and mentioned future fields of study and open ended problems. Through our brief but comprehensive survey, we aim to guide new researchers and scholar into the latest topics in Machine Translation and suggest possible areas of development.

\end{document}